\documentclass{article}
\usepackage{ijcai26}

\usepackage{times}
\usepackage{soul}
\usepackage{url}
\usepackage[hidelinks]{hyperref}
\usepackage[utf8]{inputenc}
\usepackage[small]{caption}
\usepackage{graphicx}
\usepackage{amsmath}
\usepackage{amsthm}
\usepackage{amsfonts}
\usepackage{booktabs}
\usepackage{algorithm}
\usepackage{algorithmic}
\usepackage[switch]{lineno}
\usepackage{bm}
\usepackage{multirow}
\newcommand{\best}[1]{\textbf{#1}}
\newcommand{\second}[1]{\underline{#1}}

\title{DC-VLAQ: Query--Residual Aggregation for Robust Visual Place Recognition}

\author{
Hanyu Zhu$^{1*}$\and
Zhihao Zhan$^{2}$\thanks{Equal contribution.}\and
Yuhang Ming$^{1*}$\thanks{Corresponding author.}\and
Liang Li$^3$\and
Dibo Hou$^3$\and \\
Javier Civera$^4$\And
Wanzeng Kong$^1$\\
\affiliations
$^1$BCCITA Provincial Key Laboratory, Hangzhou Dianzi University, China;
$^2$TopXGun Robotics, China; \\
$^3$ICT State Key Laboratory, Zhejiang University, China;
$^4$I3A, University of Zaragoza, Spain.
\emails
\{hanyu.zhu, yuhang.ming, kongwanzeng\}@hdu.edu.cn,
zhzhan@topxgun.com; \\
\{liang.li, houdb\}@zju.edu.cn; jcivera@unizar.es.
}

\begin{document}

\maketitle


\begin{abstract}
    One of the central challenges in visual place recognition (VPR) is learning a robust global representation that remains discriminative under large viewpoint changes, illumination variations, and severe domain shifts. 
    While visual foundation models (VFMs) provide strong local features, most existing methods rely on a single model, overlooking the complementary cues offered by different VFMs. However, exploiting such complementary information inevitably alters token distributions, which challenges the stability of existing query–based global aggregation schemes.
    To address these challenges, we propose \textbf{DC-VLAQ}, a representation-centric framework that integrates the fusion of complementary VFMs and robust global aggregation. 
    Specifically, we first introduce a lightweight \emph{residual-guided complementary fusion} that anchors representations in the DINOv2 feature space while injecting complementary semantics from CLIP through a learned residual correction. In addition, we propose the \emph{Vector of Local Aggregated Queries (VLAQ)}, a query--residual global aggregation scheme that encodes local tokens by their residual responses to learnable queries, resulting in improved stability and the preservation of fine-grained discriminative cues. 
    Extensive experiments on standard VPR benchmarks, including Pitts30k, Tokyo24/7, MSLS, Nordland, SPED, and AmsterTime, demonstrate that DC-VLAQ consistently outperforms strong baselines and achieves state-of-the-art performance, particularly under challenging domain shifts and long-term appearance changes.
\end{abstract}

\section{Introduction}
\label{sec::intro}


\begin{figure}[t]
    \centering
    \includegraphics[width=\linewidth]{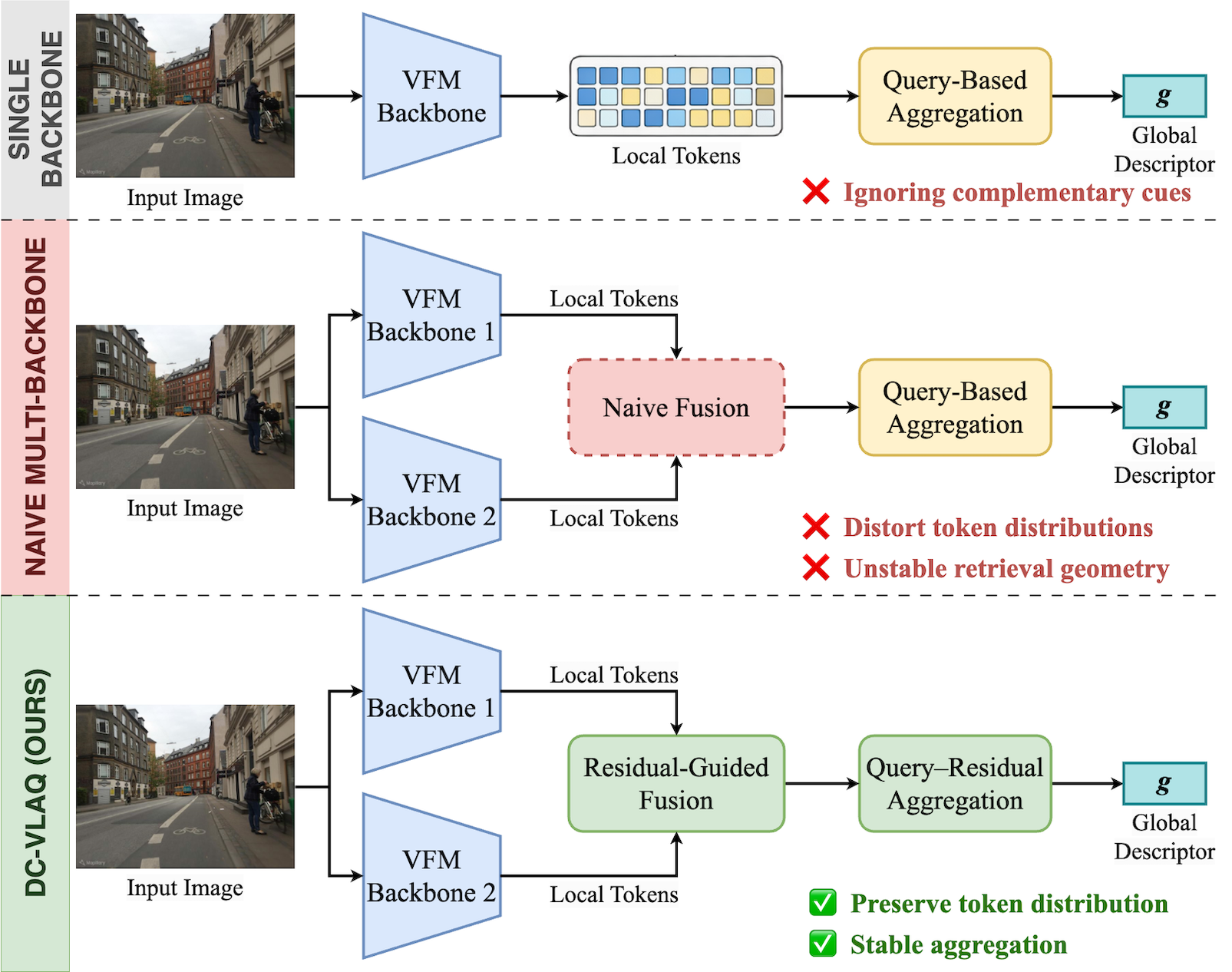}
    \caption{\textbf{Comparison of VPR pipelines:} (Top) Most existing VPR methods rely on a single VFM to extract local tokens, which ignores complementary cues across different models. (Mid.) Naive fusion of heterogeneous VFM tokens results in degraded performance, due to distorted distributions and unstable retrieval geometry. (Bot.) Our DC-VLAQ introduces \emph{Residual-Guided Complementary Fusion} to preserve the original token distribution while injecting complementary information, and \emph{Query--Residual Global Aggregation} to achieve stable and discriminative global descriptors.}
    \label{fig:teaser}
\end{figure}

Visual place recognition (VPR) aims to determine whether a visual observation corresponds to a previously visited location. 
As a core component of autonomous and mixed reality systems, VPR enables loop closure and global relocalization, which are essential for robust long-term operations. 
Despite its importance, VPR remains challenging due to severe appearance variations caused by drastic viewpoint changes, illumination differences, and domain shifts, as well as perceptual ambiguities among visually similar yet geographically distinct places~\cite{ijcai2021vprsurvey}. 
These challenges are further amplified in large-scale settings, where the diversity of the environment and increasing number of candidate locations place higher demands on the robustness and discriminability of global place representations~\cite{lindenberger2025scaling}.

Early VPR methods typically relied on handcrafted local features~\cite{sift,rublee2011orb} and global aggregators~\cite{sivic2003video,bow,vlad}.
While efficient and interpretable, these approaches have limited representation capacity and generalization ability, and therefore operate reliably only under constrained conditions. 
With the advance of deep learning, convolutional neural network (CNN)–based methods have become the dominant paradigm for VPR. Approaches such as NetVLAD~\cite{arandjelovic2016netvlad}, GeM~\cite{radenovic2018fine} and MixVPR~\cite{ali2023mixvpr} learn data-driven global descriptors that substantially improve performance through end-to-end training. 
Building upon this paradigm, subsequent works adapt learned global representations to various specific scenarios, including large-scale outdoor environments~\cite{hausler2021patch,berton2023eigenplaces}, indoor settings~\cite{cgis,aegis}, and long-term adaptation via continual learning~\cite{gao2022airloop,viper}.
However, CNN-based VPR methods still exhibit limited generalization beyond their training distributions, and often struggle to transfer reliably across different environments.

More recently, transformers~\cite{vaswani2017attention} have emerged as a promising architectural paradigm for many tasks~\cite{chen2021crossvit,zhan2025rethinking} and have been explored in VPR, with early works such as TransVPR~\cite{wang2022transvpr}. 
Building upon the this architecture, visual foundation models (VFMs) leverage large-scale pretraining to learn general-purpose visual representations with improved invariance and discriminability. Notably, AnyLoc~\cite{anyloc} demonstrates that competitive VPR performance can be achieved by directly applying frozen, pretrained VFM features without task-specific fine-tuning, highlighting the potential of foundation representations for robust place recognition.
Motivated by these findings, subsequent works further adapt and fine-tune VFM backbones for VPR tasks~\cite{lu2024towards,ali2024boq,izquierdo2024optimal}, achieving improved robustness and generalization across diverse environments and domains.

Despite the recent progress in the VFM-based VPR methods, existing methods still rely on a single backbone for local feature extraction. However, different VFMs emphasize complementary cues---for instance, fine-grained appearance cues in DINO~\cite{oquab_dinov2_2024_arxiv}, semantic concepts in CLIP~\cite{CLIP}, and geometric structure in VGGT~\cite{wang2025vggt}---suggesting clear potential for multi-model fusion. 
In practice, naively combining heterogeneous VFM features, such as direct addition, often leads to limited gains or even performance degradation.
This is primarily due to mismatched embedding spaces, which distort token distributions and induce instability in the retrieval geometry, ultimately undermining the global aggregation.

To address this, we propose \textbf{DC-VLAQ}, a representation-centric framework that jointly designs complementary feature fusion and robust global aggregation, as shown in Figure~\ref{fig:teaser}. In particular, we introduce \emph{residual-guided complementary fusion}, 
which anchors the representation in the DINO feature space and incorporates CLIP features through a learned residual correction, preserving the original token distribution while injecting complementary semantic information.
Inspired by the evolution from BoW to VLAD, we further present 
\emph{Vector of Local Aggregated Queries (VLAQ)}, a \emph{query--residual global aggregation} that encodes local tokens via their residual responses to learnable queries. This design mitigates the sensitivity of query-based aggregation to distribution shifts induced by multi-backbone fusion.

\section{Related Work}
\label{sec::related}

\paragraph{Early End-to-End VPR:}
End-to-end learning has become a dominant paradigm in VPR, with NetVLAD~\cite{arandjelovic2016netvlad} serving as a seminal milestone. 
By integrating CNN-based feature extraction with a differentiable VLAD layer, NetVLAD enables end-to-end optimization and 
achieves significant improvements over earlier handcrafted pipelines. 
This design has inspired 
a series of subsequent works like SFRS~\cite{ge2020self}, Patch-NetVLAD~\cite{hausler2021patch}, and Gated NetVLAD~\cite{gatedvlad}.
Beyond VLAD-style aggregation, 
pooling-based approaches offer a simpler alternative while still yielding effective global descriptors, as demonstrated in GeM~\cite{radenovic2018fine}.
Building on this idea, CosPlace~\cite{berton2022rethinking} and EigenPlaces~\cite{berton2023eigenplaces} both mitigate viewpoint sensitivity by reformulating the supervision at place level.

Additionally, recent works have also explored learning-based global descriptor construction through explicit feature interaction. 
MixVPR~\cite{ali2023mixvpr} introduces a lightweight MLP-based feature mixing mechanism to enhance global representations, while 
TransVPR~\cite{wang2022transvpr} incorporates a transformer-based attention on top of CNN backbones to model interactions between local and global features.
These approaches demonstrate the benefits of learnable aggregation mechanisms, but remain trained in an end-to-end manner under a single-backbone assumption.

Due to the rapid development of the field, we only review a subset of representative works here, and refer readers to recent surveys for a more comprehensive review~\cite{ijcai2021vprsurvey,yin2025general}.

\paragraph{Large-Scale Pretrained VFMs:}
Recent VFMs have demonstrated strong transferability across a wide range of vision tasks through large-scale pretraining. 
However, rather than providing a unified representation, different VFMs emphasize distinct visual cues and inductive biases, resulting in representations with complementary strengths.

Self-supervised appearance-focused models such as DINO~\cite{oquab_dinov2_2024_arxiv} learn dense patch-level representations that capture instance-level visual regularities through view-consistent training. 
Beyond appearance-level representations, other VFMs encode complementary semantic or structural priors. 
Multimodal models such as CLIP~\cite{CLIP} and SigLIP~\cite{zhai2023sigmoid} learn semantically aligned embedding spaces through large-scale image--text pretraining, exhibiting strong robustness to domain shifts and appearance variations. 
Geometry-aware VFMs, including VGGT~\cite{wang2025vggt} as well as metric depth estimation models such as MoGe2~\cite{wang2025moge2}, explicitly encode geometric structure and metric cues from visual observations. 
At the region level, object-centric foundation models such as SAM2~\cite{ravi2024sam2} and SEEM~\cite{zou2023seem} are trained with large-scale segmentation objectives, which are less aligned with dense token-based global descriptor aggregation in VPR.

\begin{figure*}[t]
    \centering
    \includegraphics[width=\textwidth]{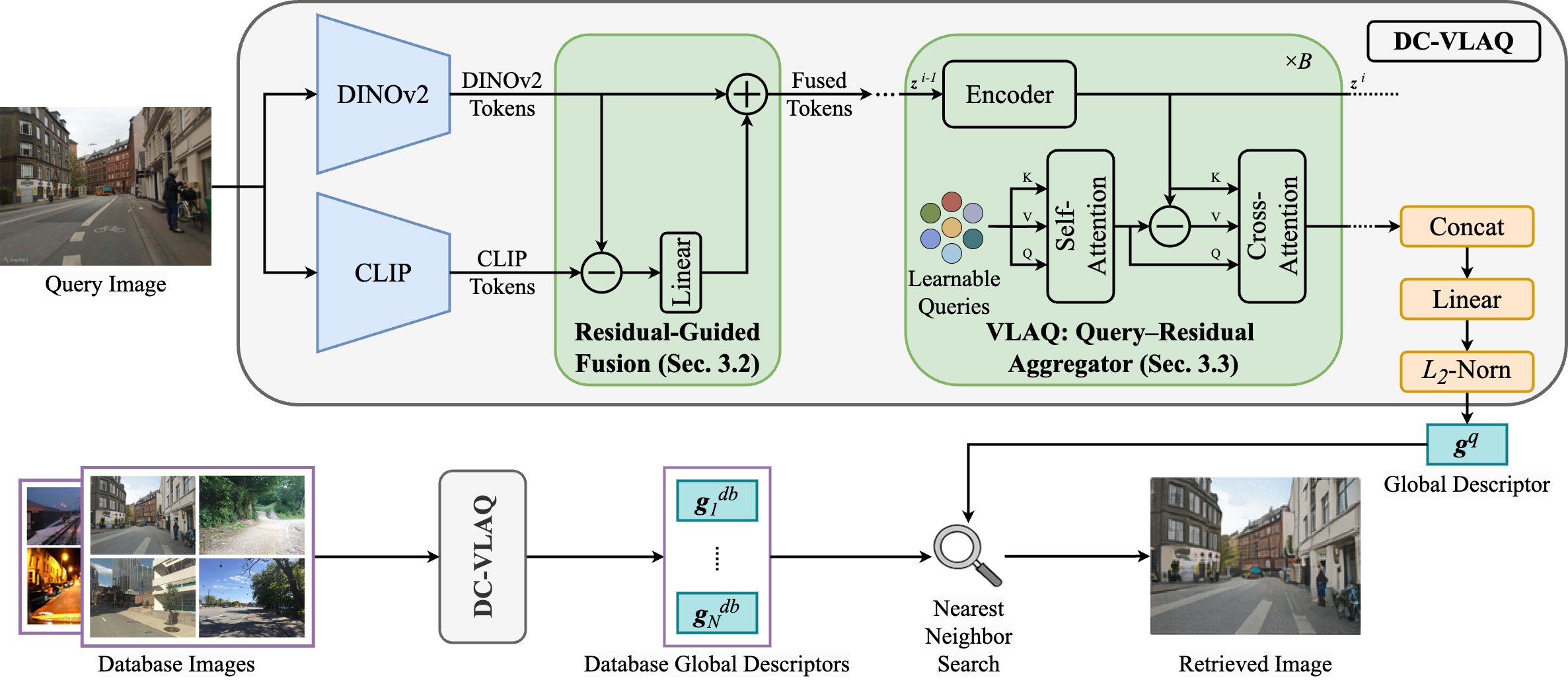}
    \caption{\textbf{Overview of the proposed DC-VLAQ pipeline.} An input image is first encoded by DINOv2 and CLIP to extract complementary local features. Then, a residual-guided fusion module injects semantic information from CLIP as residual corrections anchored to DINOv2 features. Finally, the fused tokens are aggregated by the proposed VLAQ aggregator to produce a compact global descriptor for nearest-neighbor place retrieval.}
    \vspace{-1ex}
    \label{fig:architecutre}
\end{figure*}

\paragraph{Modern VFM-based VPR:}
VFMs have reshaped VPR by enabling strong, transferable representations pretrained at scale. This is first highlighted by AnyLoc~\cite{anyloc}, which demonstrated that competitive performance across diverse environments can be achieved by directly leveraging frozen DINOv2~\cite{oquab_dinov2_2024_arxiv} features.

Building on this insight, subsequent works explore how to adapt and refine VFM representations for VPR. 
Some methods focus on backbone adaptation through fine-tuning or parameter-efficient training. 
For example, DINO-MIX~\cite{DINOMIX} fine-tunes only the last few blocks of DINOv2, while CricaVPR~\cite{lu2024cricavpr} and SelaVPR~\cite{lu2024towards} adopt parameter-efficient fine-tuning strategies, freezing most of the backbone and introducing lightweight adapters. 
Beyond backbone adaptation, several works focus on improving local feature quality and selection.
EDTFormer~\cite{jin2025edtformer} employs a lightweight decoder Transformer with low-rank adaptation to refine DINOv2 features, while FoL~\cite{wang2025focus} enhances retrieval by identifying reliable and discriminative local regions within VFM feature maps.

Another line of research focuses on constructing compact and structured global descriptors from VFM tokens. 
SALAD~\cite{izquierdo2024optimal} extends the VLAD formulation by introducing optimal transport to compute soft token-to-cluster assignments, improving the robustness of residual-based aggregation.
CliqueMining~\cite{izquierdo2024close} further improves this framework through a task-aware batch sampling strategy that emphasizes visually similar, short-range locations near decision boundaries, encouraging finer spatial discrimination.
In parallel, BoQ~\cite{ali2024boq} proposes a bag-of-queries representation that aggregates VFM tokens via learnable queries, yielding compact global descriptors while preserving implicit semantic-aware patterns.

\section{Methodology}
\label{sec::method}

\subsection{Problem Formulation}

Following most previous works, \emph{e.g.},~\cite{ali2023mixvpr,ali2024boq,izquierdo2024optimal}, we formulate VPR as a retrieval problem over a database of images of previously visited places. 
Given a target environment or map, we create a reference database $\mathcal{D} = \{\bm{I}_i\}_{i=1}^N$, where each image $\bm{I}_i$ corresponds to a distinct place observed during prior traversal.
At test time, the goal of VPR is, for a query image $\bm{I}_q$, to identify its most similar database place in the learned embedding space.

Formally, each image $\bm{I}_i$ is first processed by a visual encoder $\mathcal{E}(\cdot)$, which extracts a set of local feature embeddings $\bm{X}_i$ of dimension $d_f$:
\begin{equation}
    \bm{X}_i = \mathcal{E}(\bm{I}_i) = \{\bm{x}_{ij}\}_{j=1}^M, \; \bm{x}_{ij} \in \mathbb{R}^{d_f}.
\end{equation}
A global aggregation function $\mathcal{A}(\cdot)$ then maps the unordered set of local features $\bm{X}_i$ into a compact global descriptor $\bm{g}_i$:
\begin{equation}
    \bm{g}_i = \mathcal{A}(\bm{X}_i), \; \bm{g}_i \in \mathbb{R}^{d_g}.
\end{equation}
Given the query descriptor $\bm{g}_q = \mathcal{A}\left(\mathcal{E}\left(\bm{I}_q\right)\right)$ and the descriptors $\mathcal{G} =\{\bm{g}_i\}_{i=1}^N$ for the images in $\mathcal{D}$, retrieval is performed by nearest-neighbor search in the descriptor space using a similarity or distance function.

\subsection{Residual-Guided Complementary Fusion}

An overview of our model is shown in Figure~\ref{fig:architecutre}. Given an input image $\bm{I}_i$, we extract local features using two complementary visual encoders: DINOv2~\cite{oquab_dinov2_2024_arxiv}, and the visual encoder of CLIP~\cite{CLIP}. Formally, token-level feature representations can be obtained as: 
\begin{equation}
    \bm{X}_i^D = \mathcal{E}_{DINO}(\bm{I}_i), \; \bm{X}_i^C = \mathcal{E}_{CLIP}(\bm{I}_i),
\end{equation}
where $\bm{X}_i^D, \bm{X}_i^C \in \mathbb{R}^{M\times d}$ denote sets of local features with $M$ tokens and feature dimension $d$.
DINOv2 features provide stable appearance-driven visual cues (e.g., texture, edges, layout patterns) that are effective for retrieval, while CLIP features encode high-level semantic priors that have a coarser spatial localization but show higher invariance to drastic appearance changes.

A key challenge in multi-backbone VPR is that features from different encoders reside in incompatible embedding spaces with distinct inductive biases. 
Naively combining such features--\emph{e.g.}, by direct addition or cross-attention--often disrupts the original feature distribution, resulting in unstable retrieval geometry and unreliable nearest-neighbor relationships. 
This issue is further amplified when global aggregation is applied on top of heterogeneous feature distributions, as consistently observed in our ablation studies on backbone selection and fusion strategies (Tables~\ref{tab:ablation_backbone},~\ref{tab:ablation_fusion}).

To address this challenge, we adopt an asymmetric, anchor-based fusion strategy. 
Rather than forcing both encoders into a newly learned joint space, we treat the DINOv2 feature space as a stable appearance base manifold, and inject complementary semantic information from CLIP in the form of residual corrections.
Concretely, each fused token is constructed as
\begin{equation}
\bm{Z}_i = \bm{X}^{D}_i + F_C\left(\bm{X}^{C}_i - \bm{X}^{D}_i\right),
\label{eq:residual_fusion}
\end{equation}
where $F_C(\cdot)$ denotes a linear layer applied token-wise. 
The difference term $\mathbf{X}^{C}_i - \mathbf{X}^{D}_i$ captures the complementary information provided by CLIP relative to the DINOv2 representation, while the residual formulation constrains such deviations to remain local, ensuring that the fused features stay anchored to the original retrieval geometry.

By preserving DINOv2 as the backbone representation and modulating it only where complementary information is beneficial, this residual-guided fusion stabilizes the token distribution and mitigates embedding-space conflicts induced by naive multi-backbone fusion.
The resulting fused token set $\mathcal{Z}_i = \{\bm{z}_{ij}\}_{j=1}^M$ forms a coherent and retrieval-friendly local representation, which is subsequently fed into the global aggregation module described in Section 3.3.

\subsection{Query--Residual Global Aggregation}

Given the fused local features 
$\mathcal{Z}_i = \{\bm{z}_{ij}\}$ for image $\bm{I}_i$, our goal is to construct a compact global descriptor that remains stable for retrieval under multi-backbone fusion.
A key challenge arises from the use of absolute query-based aggregation: directly accumulating token responses based on similarity to query centers is sensitive to feature scale, response imbalance, and distribution shifts, which are particularly detrimental in retrieval-based VPR.

To address this, we propose the \emph{Vector of Local Aggregated Queries} (VLAQ), which introduces residual aggregation into a query-based pooling framework. 
Instead of aggregating absolute query responses, VLAQ encodes local features relative to their associated query prototypes, capturing how tokens deviate from reference queries.
This residual formulation mitigates sensitivity to feature magnitude and response imbalance, leading to more stable and discriminative global representations.

Concretely, we adopt a query-based assignment mechanism following BoQ~\cite{ali2024boq}, 
where $B$ blocks of learnable queries are used, each containing $S$ query vectors $\{\bm{q}_k\}_{k=1}^S$ to
softly partitions the local feature space. 
For each token–query pair, we compute a scaled dot-product similarity
\begin{equation}
    s_{ijk} = \frac{\bm{q}_k^T\bm{z}_{ij}}{\sqrt{d}},
\end{equation}
and obtain soft assignment weights via
\begin{equation}
    \alpha_{ijk} = \text{Softmax}_j(s_{ijk}).
\end{equation}

Unlike BoQ, which aggregates query responses directly, our proposed VLAQ performs residual aggregation with respect to the query centers:
\begin{equation}
    \bm{v}_{ik}=\sum_j\alpha_{ijk}(\bm{z}_{ij}-\bm{q}_k).
\end{equation}

This formulation parallels the evolution from BoW~\cite{sivic2003video} to VLAD~\cite{vlad} with handcrafted features,
transforming query-based pooling from a soft counting mechanism into a discriminative residual encoding that preserves relative structure in the feature space.

The aggregated residuals $\{\bm{v}_{ik} \}$ are concatenated $\bar{\bm{g}}_{i} = \begin{bmatrix}
    \bm{v}_{i1}^\top & \hdots & \bm{v}_{iS}^\top
\end{bmatrix}^\top$ and normalized ${\bm{g}}_{i} = \frac{\bar{\bm{g}}_{i}}{||\bar{\bm{g}}_{i}||}$ to form the final global descriptor ${\bm{g}}_{i}$ for image $\bm{I}_i$. Compared to the original BoQ formulation, VLAQ exhibits improved robustness to response imbalance and distribution shifts, while retaining the efficiency and interpretability of query-based aggregation. Alternative assignment strategies can be integrated within this residual framework and are analyzed in the ablation study.

\section{Experiments}
\label{sec::exp}

\begin{table*}[t]
\centering
\caption{\textbf{Comparison against baselines on VPR benchmark datasets.} The best is highlighted in \textbf{bold} and the second best is \underline{underlined}. 
$^\dagger$~CricaVPR uses a cross-image encoder to correlate multiple images per place and is therefore excluded from the Pitts30k comparison. 
$^\ddagger$~SALAD-CM leverages MSLS as an additional training set and is thus excluded from the MSLS comparison.}
\label{tab:sota_vpr}
\vspace{-1ex}

\setlength{\tabcolsep}{3.2pt} 
\renewcommand{\arraystretch}{1.12}

\resizebox{\textwidth}{!}{%
\begin{tabular}{l|l|ccc|ccc|ccc|ccc|ccc}
\hline
\multirow{2}{*}{Method} &
Venue & 
\multicolumn{3}{c|}{Pitts30k-test} &
\multicolumn{3}{c|}{Tokyo24/7} &
\multicolumn{3}{c|}{MSLS-val} &
\multicolumn{3}{c|}{MSLS-challenge} &
\multicolumn{3}{c}{Nordland} \\
\cline{3-17}
& \& Year
& R@1 & R@5 & R@10
& R@1 & R@5 & R@10
& R@1 & R@5 & R@10
& R@1 & R@5 & R@10
& R@1 & R@5 & R@10 \\
\hline
NetVLAD & \footnotesize{CVPR'16} &
81.9 & 91.2 & 93.7 &
60.6 & 68.9 & 74.6 &
53.1 & 66.5 & 71.1 &
35.1 & 47.4 & 51.7 &
6.4 & 10.1 & 12.5 \\

SFRS & \footnotesize{ECCV'20} &
89.4 & 94.7 & 95.9 &
81.0 & 88.3 & 92.4 &
69.2 & 80.3 & 83.1 &
41.6 & 52.0 & 56.3 &
16.1 & 23.9 & 28.4 \\

Patch-NetVLAD & \footnotesize{CVPR'21} &
88.7 & 94.5 & 95.9 &
86.0 & 88.6 & 90.5 &
79.5 & 86.2 & 87.7 &
48.1 & 57.6 & 60.5 &
44.9 & 50.2 & 52.2 \\

TransVPR & \footnotesize{CVPR'22} &
89.0 & 94.9 & 96.2 &
79.0 & 82.2 & 85.1 &
86.8 & 91.2 & 92.4 &
63.9 & 74.0 & 77.5 &
63.5 & 68.5 & 70.2 \\

CosPlace & \footnotesize{CVPR'22} &
88.4 & 94.5 & 95.7 &
81.9 & 90.2 & 92.7 &
82.8 & 89.7 & 92.0 &
61.4 & 72.0 & 76.6 &
58.5 & 73.7 & 79.4 \\

EigenPlaces & \footnotesize{ICCV'23} &
92.5 & 96.8 & 97.6 &
93.0 & 96.2 & 97.5 &
89.1 & 93.8 & 95.0 &
67.4 & 77.1 & 81.7 &
71.2 & 83.8 & 88.1 \\

MixVPR & \footnotesize{WACV'23} &
91.5 & 95.5 & 96.3 &
85.1 & 91.7 & 94.3 &
88.0 & 92.7 & 94.6 &
64.0 & 75.9 & 80.6 &
76.2 & 86.9 & 90.3 \\

SelaVPR & \footnotesize{ICLR'24} &
92.8 & 96.8 & 97.7 &
94.0 & 96.8 & 97.5 &
90.8 & 96.4 & 97.2 &
73.5 & 87.5 & 90.6 &
87.3 & 93.8 & 95.6 \\

CricaVPR$^\dagger$ & \footnotesize{CVPR'24} &
{94.9}$^\dagger$ & {97.3}$^\dagger$ & {98.2}$^\dagger$ &
93.0 & 97.5 & 98.1 &
90.0 & 95.4 & 96.4 &
69.0 & 82.1 & 85.7 &
\second{90.7} & 96.3 & \second{97.6} \\

BoQ & \footnotesize{CVPR'24} &
\second{93.7} & \second{97.1} & \second{97.9} &
\second{98.1} & \second{98.1} & \second{98.7} &
\second{93.8} & 96.8 & 97.0 &
\second{79.0} & 90.3 & 92.0 &
90.6 & 96.0 & 97.5 \\

SALAD & \footnotesize{CVPR'24} &
92.5 & 96.4 & 97.5 &
94.6 & 97.5 & 97.8 &
92.2 & 96.4 & 97.0 &
75.0 & 88.8 & 91.3 &
89.7 & 95.5 & 97.0 \\

SALAD-CM$^\ddagger$ & \footnotesize{ECCV'24} &
92.7 & 96.8 & \second{97.9} &
94.6 & 97.5 & 97.8 &
94.2$^\ddagger$ & 97.2$^\ddagger$ & 97.4$^\ddagger$ &
82.7$^\ddagger$ & 91.2$^\ddagger$ & 92.7$^\ddagger$ &
\second{90.7} & \second{96.6} & 97.5 \\



EDTFormer & \footnotesize{TCSVT'25} &
93.4 & 97.0 & \second{97.9} &
97.1 & 98.1 & 98.4 &
92.0 & 96.6 & 97.2 &
78.4 & 89.8 & 91.9 &
88.3 & 95.3 & 97.0 \\

FoL-global & \footnotesize{AAAI'25} &
- & - & - &
96.2 & 98.7 & 98.7 &
93.1 & \second{96.9} & \second{97.4} &
78.7 & \second{90.8} & \second{93.0} &
87.8 & - & - \\

\hline

DC-VLAQ (Ours) & - &
\best{94.3} & \best{97.6} & \best{98.3}&
\best{98.7} & \best{99.7} & \best{99.7} &
\best{94.2} & \best{97.3} & \best{97.6} &
\best{81.7} & \best{92.2} & \best{94.5} &
\best{92.8} & \best{97.2} & \best{98.2} \\
\hline
\end{tabular}%
}
\end{table*}

\begin{table}[t]
\centering
\caption{\textbf{Comparison against baselines on robustness-oriented VPR benchmarks.}
The best results are shown in \textbf{bold} and the second best are \underline{underlined}.}
\label{tab:robust_vpr}
\vspace{-1ex}

\setlength{\tabcolsep}{4pt}
\renewcommand{\arraystretch}{1.15}

\resizebox{1.05\linewidth}{!}{%
\begin{tabular}{l|l|ccc|ccc}
\hline
\multirow{2}{*}{Method} &
Venue &
\multicolumn{3}{c|}{SPED} &
\multicolumn{3}{c}{AmsterTime} \\
\cline{3-8}
& \& Year
& R@1 & R@5 & R@10
& R@1 & R@5 & R@10 \\
\hline

NetVLAD & \footnotesize{CVPR'16} &
78.7 & 88.3 & 91.4 &
16.3 & -- & -- \\

GeM & \footnotesize{TPAMI'18} &
64.6 & 79.4 & 83.5 &
-- & -- & -- \\




CosPlace & \footnotesize{CVPR'22} &
75.3 & 85.9 & 88.6 &
47.7 & -- & -- \\

EigenPlaces & \footnotesize{ICCV'23} &
82.4 & 91.4 & {94.7} &
48.9 & -- & -- \\

MixVPR & \footnotesize{WACV'23} &
{85.2} & {92.1} & 94.6 &
40.2 & -- & -- \\

SelaVPR & \footnotesize{ICLR'24} &
89.5 & - & - &
-- & -- & -- \\

CricaVPR & \footnotesize{CVPR'24} &
-- & -- & -- &
64.7 & 82.8 & 87.5 \\

BoQ & \footnotesize{CVPR'24} &
\second{92.5} & \second{95.9} & \second{96.7} &
{63.0} & {81.6} & {85.1} \\

SALAD & \footnotesize{CVPR'24} &
92.1 & 96.2 & -- &
-- & -- & -- \\

SALAD-CM & \footnotesize{ECCV'24} &
89.3 & -- & -- &
-- & -- & -- \\

EDTFormer & \footnotesize{TCSVT'25} &
-- & -- & -- &
\second{65.2} & \second{85.0} & \best{89.0} \\

FoL-global & \footnotesize{AAAI'25} &
92.1 & -- & -- &
{64.6} & -- & -- \\

\hline

DC-VLAQ (Ours) & - &
\best{93.9} & \best{97.7} & \best{98.2} &
\best{66.8} & \best{85.6} & \second{88.9} \\
\hline

\end{tabular}
}
\end{table}

\begin{figure*}[t]
    \centering
    \includegraphics[width=\textwidth]{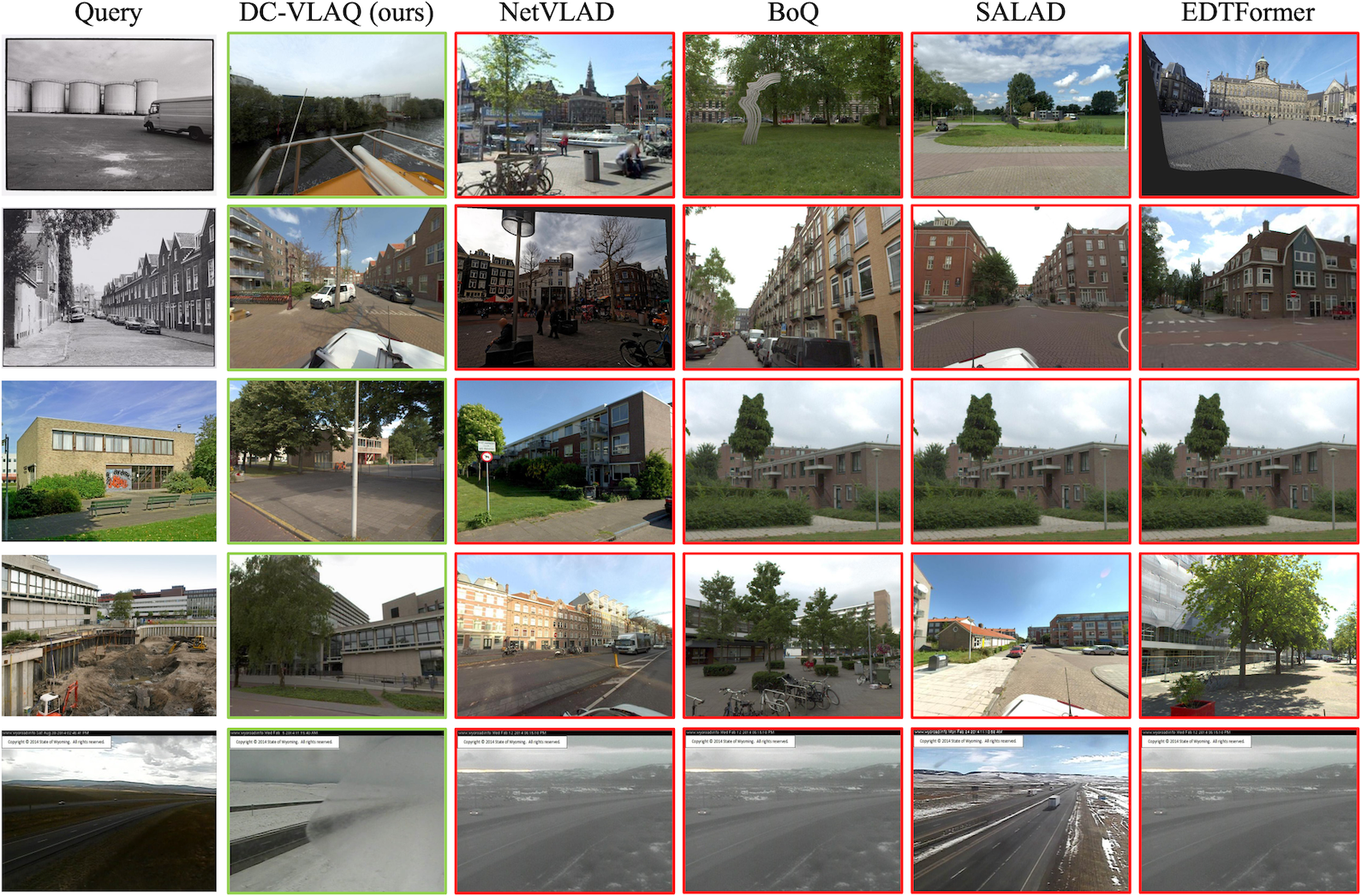}
    \caption{\textbf{Qualitative comparison of retrieval results on challenging VPR benchmarks.}
    The first four rows show results on AmsterTime, characterized by severe historical domain shifts across years, while the last row reports results on SPED, featuring cross-domain surveillance imagery and extreme appearance changes.
    For each query (leftmost), we visualize the top-1 retrieved images produced by different methods, with correct retrievals highlighted in green and incorrect ones in red.
    }
    \vspace{-1ex}
    \label{fig:results}
\end{figure*}

\subsection{Datasets and Evaluation Metrics}

Following the common methodology for evaluation of VPR pipelines~\cite{ali2024boq,izquierdo2024optimal,izquierdo2024close}, we use 
\textit{GSV-Cities}~\cite{ali2022gsv} as the training dataset and conduct evaluation on
\textit{Pitts30k}~\cite{torii2013visual}, 
\textit{Tokyo24/7}~\cite{torii201524}, 
\textit{MSLS}~\cite{warburg2020mapillary}, 
\textit{Nordland}~\cite{olid2018single}, 
\textit{SPED}~\cite{zaffar2021sped}, and 
\textit{AmsterTime}~\cite{yildiz2022amstertime}. 
\textit{GSV-Cities} 
is a large-scale street-view dataset collected across diverse cities worldwide. It provides rich visual diversity and covers common VPR challenges, including viewpoint changes, illumination variations, seasonal appearance shifts, and long-term domain changes, making it well suited for learning robust and generalizable global place representations.

\textit{Pitts30k} serves as a standard large-scale urban benchmark, primarily evaluating viewpoint robustness under complex architectural environments.
Also in urban environments, \textit{Tokyo24/7} emphasizes illumination robustness, particularly under challenging day--night transitions. 
\textit{MSLS} extends to suburban and rural environments, introducing strong temporal appearance variation across different acquisition times and devices.
Additionally, \textit{Nordland} provides a complementary perspective, focusing on severe seasonal change along a fixed traversal to isolate temporal robustness.
In contrast, \textit{SPED} consists of cross-domain images captured by surveillance cameras, offering dense long-term observations, 
and \textit{AmsterTime} targets extreme historical domain shift, where queries and references are captured years apart with different devices and substantially altered visual characteristics.

For all evaluation datasets, we adopt Recall@K (R@1, R@5, and R@10) as the evaluation metric. A query is considered correctly localized if at least one of the top-$K$ retrieved reference images corresponds to a ground-truth positive according to the dataset protocol. Recall@K is widely used in VPR and provides a consistent measure of retrieval performance across different benchmarks.

\subsection{Implementation Details}

We adopt pretrained DINOv2 and CLIP as visual feature extractors. In all experiments, we fine-tune the last two blocks of the DINOv2 encoder, while keep CLIP visual encoder frozen. 
We also apply $L2$ norm to each encoder's local feature to stabilize the fusion process.
For the global aggregator, we use 2 VLAQ blocks with 64 learnable queries each.

The training follows the standard VPR protocol~\cite{ali2022gsv} and employs the Multi-Similarity loss~\cite{Wang_2019_CVPR}.
To ensure sufficient positive and negative pairs within each batch, we use a place-balanced batching strategy with each batch containing 110 4-image places, i.e. 440 images in total.
The network is optimized using AdamW~\cite{loshchilov2018decoupled} for 40 epochs with linear warm-up in the first 10 epochs.
The initial learning rate is set to $2\times10^{-4}$ with a weight decay of $10^{-3}$, and is decayed by a factor of 0.1 every 10 epochs.
To improve fine-tuning stability, a reduced learning-rate multiplier of $0.2\times$ is applied to DINOv2 blocks.
Finally, images are resized to $280\times280$ during training, and a higher resolution of $322\times322$ for evaluation.


\subsection{Comparison Results}

\paragraph{Baselines.}
We compare our approach with a wide range of representative VPR baselines, with all results acquired from their original paper. We first set the baseline performance with the seminal VPR model NetVLAD~\cite{arandjelovic2016netvlad}. 
We then include early end-to-end VPR models including GeM~\cite{radenovic2018fine}, SFRS~\cite{ge2020self}, 
Patch-NetVLAD~\cite{hausler2021patch}, TransVPR~\cite{wang2022transvpr}, CosPlace~\cite{berton2022rethinking}, EigenPlaces~\cite{berton2023eigenplaces}, and MixVPR~\cite{ali2023mixvpr}.
We also compare against several recent state-of-the-art VFM-based methods, such as SelaVPR~\cite{lu2024towards}, CricaVPR~\cite{lu2024cricavpr}, SALAD~\cite{izquierdo2024optimal}, SALAD-CM~\cite{izquierdo2024close}, EDTFormer~\cite{jin2025edtformer}, FoL-global~\cite{wang2025focus}, as well as BoQ~\cite{ali2024boq}, which serves as a direct query-based aggregation baseline for our method.

\paragraph{Results and Analysis.}
Quantitative results on standard and robustness-oriented benchmarks are reported in Tables~\ref{tab:sota_vpr} and~\ref{tab:robust_vpr}. Across all datasets and evaluation metrics, DC-VLAQ consistently achieves the best or second-best performance, demonstrating strong effectiveness and generalization under diverse conditions. 
More importantly, DC-VLAQ exhibits particularly pronounced performance gain on Nordland and MSLS-challenge, both of which feature severe appearance variation and reduced temporal continuity. 
Along with the notable increment in SPED and AmsterTime, these results demonstrate that DC-VLAQ generalizes well under severe domain shifts across both time and sensing conditions.

\begin{figure}[t]
    \centering
    \includegraphics[width=\linewidth]{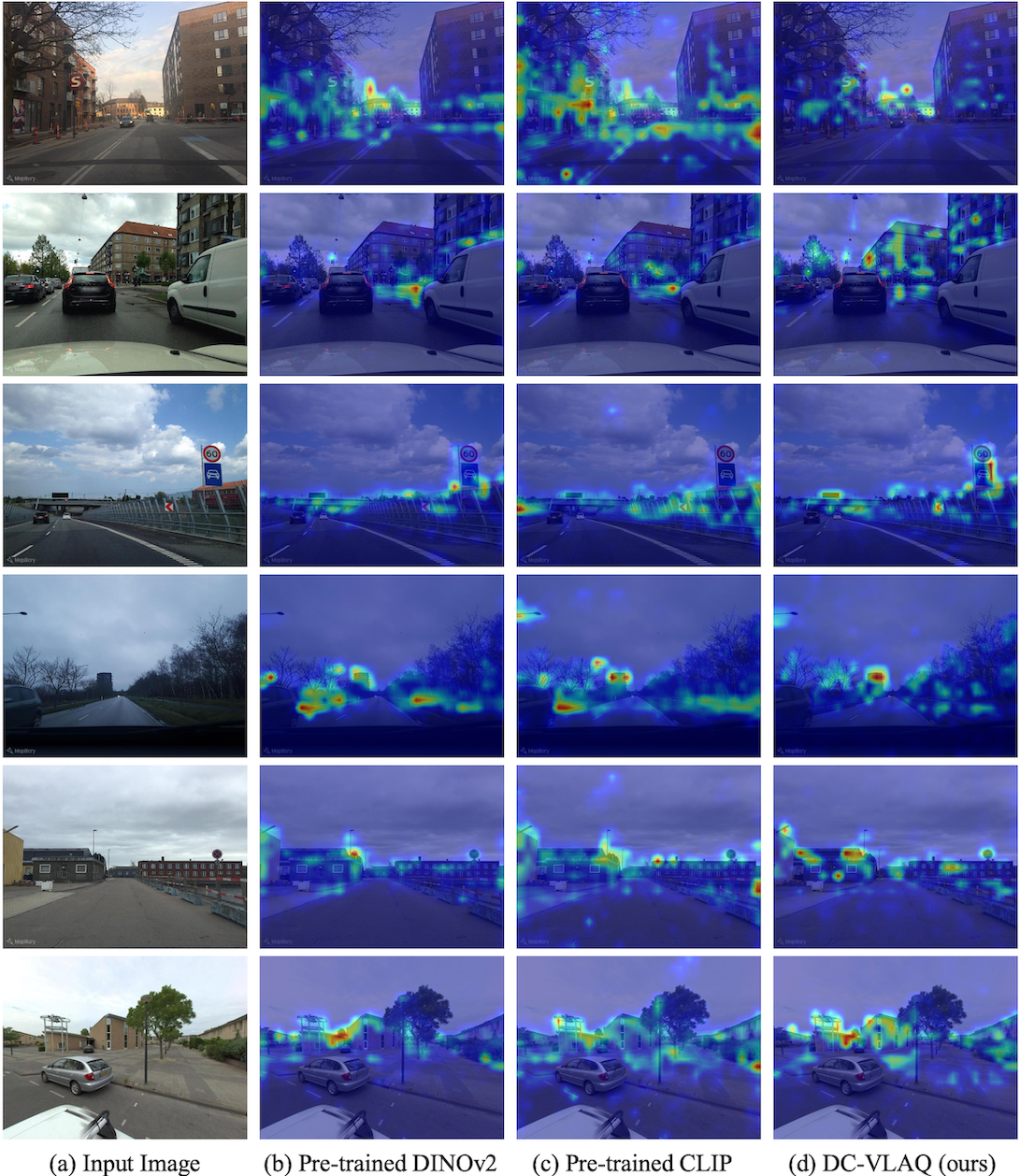}
    \caption{\textbf{Visualization of query activation heatmaps.} 
    The first five rows correspond to MSLS-val, and the last row shows examples from AmsterTime.
    For each query (a), we visualize the activation heatmaps produced by pre-trained DINOv2 (b), pre-trained CLIP (c), and our DC-VLAQ (d). Observe the complementarity of DINOv2 and CLIP, and the adaptation of our DC-VLAQ to the VPR task, \emph{e.g.}, by suppressing non-discriminative road features.}
    \label{fig:query-map}
\end{figure}

Moreover, Figure~\ref{fig:results} presents qualitative retrieval results on challenging VPR benchmarks. 
DC-VLAQ consistently retrieves visually and structurally consistent matches under challenging appearance changes, whereas baseline methods often fail due to over-reliance on global appearance cues. 
Figure~\ref{fig:query-map} further provides insight into the internal behavior of different representations by visualizing query activation heatmaps. 
Compared to pre-trained DINOv2 and CLIP, DC-VLAQ produces more focused and spatially consistent activations on stable structural elements such as building facades, road boundaries, and static landmarks, while suppressing transient or less informative regions.
This behavior aligns well with the design of DC-VLAQ, where residual-guided fusion preserves a stable appearance manifold and query–residual aggregation emphasizes relative local deviations rather than absolute responses.

\subsection{Ablation Studies}

To verify the validity of our design choice, we further conduct ablation studies on \textit{VFM backbone choices}, \textit{local feature fusion approaches}, and \textit{global aggregation strategies}. 
In each experiment, only one design choice is modified, while all the other components remain the same as our final model.

\paragraph{VFM Backbone Choices:} 
We first analyze the impact of different VFM backbones, with results reported in Table~\ref{tab:ablation_backbone}. 
Among single-backbone configurations, DINOv2 outperforms CLIP and VGGT on both MSLS-val and Pitts30k-test, validating our choice of anchoring fused feature distribution on the DINOv2 feature space.
Moreover, combining DINOv2 and CLIP yields a clear performance improvement over all single-backbone baselines. This confirms that CLIP provides complementary semantic information that can enhance DINOv2 features when integrated in a residual-guided manner. 
In contrast, incorporating VGGT, although improving over VGGT alone, leads to slight performance drop compared to DINOv2. 
This suggests that while residual-guided fusion is effective, further gains from additional VFM backbones depend on the degree of task-relevant complementarity.

\begin{table}[t]
\centering
\caption{Ablation study on different VFM backbone choices. 
}
\vspace{-1ex}
\label{tab:ablation_backbone}
\resizebox{\linewidth}{!}{
\begin{tabular}{l|ccc|ccc}
\hline
\multirow{2}{*}{Backbone} 
& \multicolumn{3}{c|}{MSLS-val} 
& \multicolumn{3}{c}{Pitts30k-test} \\
\cline{2-7}
& R@1 & R@5 & R@10 & R@1 & R@5 & R@10 \\
\hline
DINOv2                    & 93.1 & 96.8 & 97.3 & 93.7 & 96.9 & 98.0 \\
CLIP                      & 91.4 & 96.4 & 97.0 & 92.7 & 96.5 & 97.5 \\
VGGT                      & 92.2 & 95.8 & 96.4 & 92.1 & 96.1 & 97.0 \\
DINOv2 \& CLIP            & \best{94.2} & \best{97.3} & \best{97.6} & \best{94.3} & \best{97.6} & \best{98.3} \\
DINOv2 \& VGGT            & 92.6 & 96.5 & 96.8 & 93.2 & 96.8 & 97.7 \\
DINOv2 \& VGGT \& CLIP & 93.0 & 96.5 & 96.9 & 93.4 & 97.0 & 97.9 \\
\hline
\end{tabular}
}
\end{table}

\begin{table}[t]
\centering
\caption{Ablation study on local feature fusion approaches evaluated on MSLS-val and Pitts30k-test. 
}
\vspace{-1ex}
\label{tab:ablation_fusion}
\resizebox{\linewidth}{!}{
\begin{tabular}{l|ccc|ccc}
\hline
\multirow{2}{*}{Fusion Method} 
& \multicolumn{3}{c|}{MSLS-val} 
& \multicolumn{3}{c}{Pitts30k-test} \\
\cline{2-7}
& R@1 & R@5 & R@10 & R@1 & R@5 & R@10 \\
\hline
Naive Addition  & 93.0 & 96.5 & 97.2 & 93.5 & 97.2 & 98.0 \\
Cross-Attention & 93.2 & 96.4 & 97.2 & 93.5 & 97.1 & 98.0 \\
FiLM            & 93.0 & 96.5 & 96.9 & 93.3 & 97.0 & 97.9 \\
Residual (Ours) & \best{94.2} & \best{97.3} & \best{97.6} & \best{94.3} & \best{97.6} & \best{98.3} \\
\hline
\end{tabular}
}
\end{table}

\begin{table}[t!]
\centering
\caption{Ablation study on global aggregation strategies evaluated on MSLS-val and Pitts30k-test.}
\vspace{-1ex}
\label{tab:ablation_aggregation}
\resizebox{\linewidth}{!}{
\begin{tabular}{l|ccc|ccc}
\hline
\multirow{2}{*}{Aggregation} 
& \multicolumn{3}{c|}{MSLS-val} 
& \multicolumn{3}{c}{Pitts30k-test} \\
\cline{2-7}
& R@1 & R@5 & R@10 & R@1 & R@5 & R@10 \\
\hline
BoQ           & 93.4 & 96.2 & 97.3 & 93.8 & 97.1 & 98.0 \\
OT-BoQ        & 92.3 & 96.2 & 96.8 & 93.8 & 97.3 & 98.1 \\
VLAQ (Ours)   & \best{94.2} & \best{97.3} & \best{97.6} & \best{94.3} & \best{97.6} & \best{98.3} \\
\hline
\end{tabular}
}
\end{table}

\paragraph{Local Features Fusion Approaches:} 
We then summarize the results on different local feature fusion strategies for DINOv2 and CLIP features in Table~\ref{tab:ablation_fusion}. 
Naive addition directly sums the two features and serves as a simple baseline. 
We also evaluate more expressive fusion mechanisms, including cross-attention~\cite{vaswani2017attention} and FiLM~\cite{perez2018film}.
However, despite their higher modeling capacity, these methods do not yield consistent performance gains and in some cases even degrade retrieval recall.
In contrast, our residual-based fusion consistently improves performance across all benchmarks.
This suggests that directly modeling complex interactions between heterogeneous features may disturb the underlying retrieval geometry, whereas residual fusion preserves the original DINOv2 feature distribution and injects complementary semantic information in a controlled manner.

\paragraph{Global Aggregation Strategies:}
We finally study the different global aggregation strategies and present results in Table~\ref{tab:ablation_aggregation}. 
Using the BoQ aggregator~\cite{ali2024boq} yields only marginal improvements over the single-backbone DINOv2 baseline. This suggests that directly aggregating absolute query responses remains sensitive to response imbalance and distribution shifts introduced by multi-backbone fusion.
We further evaluate OT-BoQ, which incorporates optimal transport–based assignment inspired by SALAD~\cite{izquierdo2024optimal}.
While OT improves the assignment regularity, it does not lead to consistent performance gains in our setting.
In contrast, our proposed VLAQ consistently achieves clear performance improvements across datasets.
These results confirm that residual aggregation, rather than assignment refinement alone, is crucial for robust global descriptor construction under multi-backbone fusion.


\section{Conclusion}
\label{sec:conclusion}
In this paper, we proposed \textbf{DC-VLAQ}, a representation-centric framework for robust visual place recognition that jointly addresses complementary feature fusion and global descriptor aggregation. By introducing a residual-guided fusion strategy, DC-VLAQ integrates semantic cues from CLIP into a stable DINOv2 feature space without disrupting its retrieval geometry. On top of the fused representation, we further proposed a query--residual aggregation scheme that encodes local features relative to learnable query prototypes, improving robustness to response imbalance and appearance variation.
Extensive experiments on both standard and robustness-oriented VPR benchmarks demonstrate that DC-VLAQ consistently outperforms state-of-the-art methods, particularly under severe illumination changes, seasonal variations, and long-term domain shifts. Looking forward, this framework naturally opens up several promising directions, including the integration of additional complementary visual backbones, as well as extensions to other sensing modalities such as text and point clouds.

\appendix



\section*{Acknowledgments}

This research has been supported in part by the Zhejiang Provincial Natural Science Foundation of China under Grant LQN25F030015, National Natural Science Foundation of China under Grant 62401188, Open Research Project of the State Key Laboratory of Industrial Control Technology China under Grant ICT2025B41, the Spanish Government under Grants PID2021-127685NB-I00 and PID2024-155886NB-I00, and the Aragon Government under Grant T45\_23R.


\bibliographystyle{named}
\bibliography{ijcai26}

\end{document}